\documentclass[runningheads]{llncs}
\usepackage{graphicx}
\usepackage{hhline}
\usepackage{colortbl}
\usepackage{lscape} 
\usepackage{adjustbox}
\usepackage{multirow}
\usepackage{caption}
\usepackage{subcaption}
\usepackage{placeins}
\usepackage{xcolor}
\begin{document}

\title{Automatic lesion analysis for increased efficiency in outcome prediction of traumatic brain injury}

\titlerunning{Automatic lesion analysis for increased efficiency in outcome prediction}
%


\author{
Margherita Rosnati\inst{1,2,3}
\and Eyal Soreq\inst{2,3}
\and Miguel Monteiro\inst{1}
\and Lucia Li\inst{2,3}
\and \\Neil S.N. Graham\inst{2,3}
\and Karl Zimmerman\inst{2,3}
\and Carlotta Rossi\inst{4}
\and Greta Carrara\inst{4}
\and \\Guido Bertolini\inst{4}
\and David J. Sharp\inst{2,3}
\and Ben Glocker\inst{1}
}
\index{Rosnati, Margherita}
\index{Soreq, Eyal}
\index{Monteiro, Miguel}
\index{Li, Lucia}
\index{Graham, Neil S.N.}
 \index{Zimmerman, Karl}
 \index{Bertolini, Guido}
 \index{Rossi, Carlotta}
 \index{Carrara, Greta}
  \index{Sharp, David J.}
  \index{Glocker, Ben}


%
\authorrunning{M. Rosnati et al.}

%
\institute{BioMedIA Group, Department of Computing, Imperial College London, UK \and
UK Dementia Research Institute Care Research and Technology Centre, Imperial College London and the University of Surrey, Guildford \and
Department of Brain Sciences, Faculty of Medicine, Imperial College London, UK \and
Laboratorio di Epidemiologia Clinica
Dipartimento di Salute Pubblica,
Istituto di Ricerche Farmacologiche Mario Negri IRCCS, Italy
\email{margherita.rosnati12@imperial.ac.uk}}

\maketitle              
\begin{abstract}
The accurate prognosis for traumatic brain injury (TBI) patients is difficult yet essential to inform therapy, patient management, and long-term after-care. Patient characteristics such as age, motor and pupil responsiveness, hypoxia and hypotension, and radiological findings on computed tomography (CT), have been identified as important variables for TBI outcome prediction. CT is the acute imaging modality of choice in clinical practice because of its acquisition speed and widespread availability. However, this modality is mainly used for qualitative and semi-quantitative assessment, such as the  Marshall scoring system, which is prone to subjectivity and human errors. This work explores the predictive power of imaging biomarkers extracted from routinely-acquired hospital admission CT scans using a state-of-the-art, deep learning TBI lesion segmentation method. We use lesion volumes and corresponding lesion statistics as inputs for an extended TBI outcome prediction model. We compare the predictive power of our proposed features to the Marshall score, independently and when paired with classic TBI biomarkers. We find that automatically extracted quantitative CT features perform similarly or better than the Marshall score in predicting unfavourable TBI outcomes. Leveraging automatic atlas alignment, we also identify frontal extra-axial lesions as important indicators of poor outcome. Our work may contribute to a better understanding of TBI, and provides new insights into how automated neuroimaging analysis can be used to improve prognostication after TBI.
\end{abstract}
\section{Introduction}
Traumatic brain injury (TBI) is a leading cause of death in Europe \cite{tagliaferri2006systematic}. In the UK alone, 160,000 patients with TBI are admitted to hospitals annually, with an estimated yearly cost of £15 billion \cite{parsonage2016traumatic}. 
The accurate prediction of TBI outcome is still an unresolved challenge \cite{kalanuria2013early,helmrich2021development} whose resolution could improve the therapy and after-care of patients. 
Due to its acquisition speed and wide availability, computed tomography (CT) is the imaging modality of choice in clinical practice \cite{kim2011imaging} and a key component in TBI outcome prediction \cite{carter2012imaging}.
The Marshall score \cite{marshall1992diagnosis} is one of the most widely used metrics to evaluate TBI injury severity. However, it does not leverage the rich information content of CT imaging \cite{brown2019predictive} and requires a radiologist to assess the CT scan manually, which is time-consuming.
Years of acquisition of CT of TBI patients generated rich datasets, opening the possibility of automatic extraction of CT biomarkers. These could enable a deeper and broader use of imaging data, augmenting the skills of radiologists and reducing the workload, allowing them to see more patients.
Machine learning for medical imaging is a growing research field with advances in medical imaging segmentation \cite{ronneberger2015u} and classification \cite{cirecsan2013mitosis}. It can be used for fast and autonomous outcome prediction of TBI using imaging data.

This work explores the predictive power of novel TBI biomarkers computationally extracted from hospital admission CT scans. TBI lesion volumes are automatically extracted from the scans; then, lesion statistics are derived to inform the prediction of TBI outcome. We compare the discriminate power of our proposed features to the Marshall score features independently and when paired with clinical TBI biomarkers.

In particular, we make the following contributions:
\begin{itemize}
    \item \textbf{Novel machine-learning driven imaging biomarkers.} We extract interpretable measurements for TBI lesion quantification;
    \item \textbf{Human-level performance on unfavourable outcome prediction.} We reach comparable, if not superior, performance to manually extracted CT biomarkers when predicting unfavourable outcome, both in isolation and when paired with clinical TBI biomarkers;
    \item \textbf{Imaging biomarker relevance.} We show that features relating to extra-axial haemorrhage in the frontal lobe are important for the prediction of outcome, confirming previous clinical findings in a data-driven manner.
\end{itemize}

\section{Related work}
The prediction of TBI of outcome has primarily been tackled using clinical features. 
Jiang~et~al.~\cite{jiang2002early} and Majdan~et~al.~\cite{Majdan2017} used clinical features such as age and motor score to predict the patient outcome with a regression model.  Pasipanodya~et~al.~\cite{pasipanodya2022characterizing} focused on predictions for different patient subgroups.
Huie~et~al.~\cite{huie2018neurotrauma} provide an extensive review of clinical biomarkers for the prediction of unfavourable outcome in TBI. 
More recently, Bruschetta~et~al.~\cite{Bruschetta2022} and Matsuo~et~al.~\cite{matsuo2020machine} investigated using the same predictors with machine learning models, such as support vector machines and neural networks. 
Researchers also used more complex features to predict TBI outcome, such as electroencephalograms \cite{noor2020machine} and magnetic resonance imaging (MRI) \cite{graham2021axonal}.

A stream of research within TBI focuses on features extracted from CT. 
Recent work adopted neural networks for TBI lesion segmentation and midline shift quantification
\cite{monteiro2020multiclass,jain2019automatic}.

Similarly to our work, Plassard~et~al.~\cite{plassard2015revealing} and Chaganti~et~al.~\cite{chaganti2016bayesian} used multi-atlas labelling to extract radiomics from different brain regions and predict a variety of TBI end-points, yet excluding unfavourable outcome. Pease~et~al.~\cite{Pease2022} trained a convolutional neural network using CT scans and clinical TBI biomarkers to predict the TBI outcome and achieved comparable performance to IMPACT.
Unlike the biomarkers we designed, the authors extracted deep learning features, which are not interpretable by humans. In parallel to our work, Yao~et~al.~\cite{yao2020automated} trained a neural network to segment epidural haematomas and used the resulting segmentation volumes to predict patient mortality. 
Our work differs because we focused on the more challenging and clinically relevant problem of predicting TBI unfavourable outcome. TBI outcomes are measured by the patient state scale Glasgow Outcome Scale Extended (GOS-E), where a score of 4 or below defines an unfavourable outcome.
\section{Methods}
\subsubsection*{Study design}
We analysed data from the observational studies
The Collaborative REsearch on ACute Traumatic Brain Injury in intensiVe Care Medicine in Europe (CREACTIVE, NCT02004080 
), and BIOmarkers of AXonal injury after TBI (BIO-AX-TBI, NCT03534154 \cite{graham2020multicentre}), the latter being a follow-up study to CREACTIVE. From these observational studies, sub-studies collecting CT  scans recruited 1986 patients admitted to intensive care with a diagnosis of TBI between 2013 and 2019 from 21 European hospitals. CREACTIVE recruited patients admitted to a hospital due to suspected TBI, whereas for BIO-AX-TBI, the criterion was moderate-severe TBI as defined in \cite{malec2007mayo}. The studies did not define a protocol for acquisition or scanner type. Hence, the CT scans collected were a heterogeneous dataset akin to clinical scenarios.

\subsubsection*{Data cleaning}

We discarded patients for whom trauma time, hospital admission and discharge, admission CT scan, and the patient outcome measure GOS-E were not recorded. In addition, we discarded patients who had surgery before the scan, whose scan could not be automatically aligned to an atlas, and for whom one of the clinical TBI biomarkers - age, Glasgow Coma Scale motor score and  pupil responsiveness, hypoxia and hypotension - or Marshall score was missing. A flowchart of the patient selection can be found in the appendix Figure~\ref{fig:patient_flow_chart}.
Of the remaining 646 patients, 389 (60.2\%) had an unfavourable outcome, of which 225 (34.8\%) died. The median age is 56.6, and 74.9\% of the patients are male. Out of the 21 centres, we selected three centres with 59 patients as an independent replication holdout test set, and we used the remaining 18 centres with 587 patients as a training set. 

\subsubsection*{Feature extraction}

All CT scans were standardised via automatic atlas registration to obtain rigid alignment to MNI space. 
We then used the deep-learning lesion segmentation method, BLAST-CT, described in \cite{monteiro2020multiclass}. The method produces a voxel-wise classification of intraparenchymal haemorrhages (IPH), extra-axial haemorrhages (EAH), perilesional oedemas (PLO) and intraventricular haemorrhages (IVH), and reports respective Dice similarity coefficients of 65.2\%, 55.3\%, 44.8\% and 47.3\% for lesions above 1mL. 
\begin{figure}[!t]
    \centering
    \includegraphics[width=.7\textwidth]{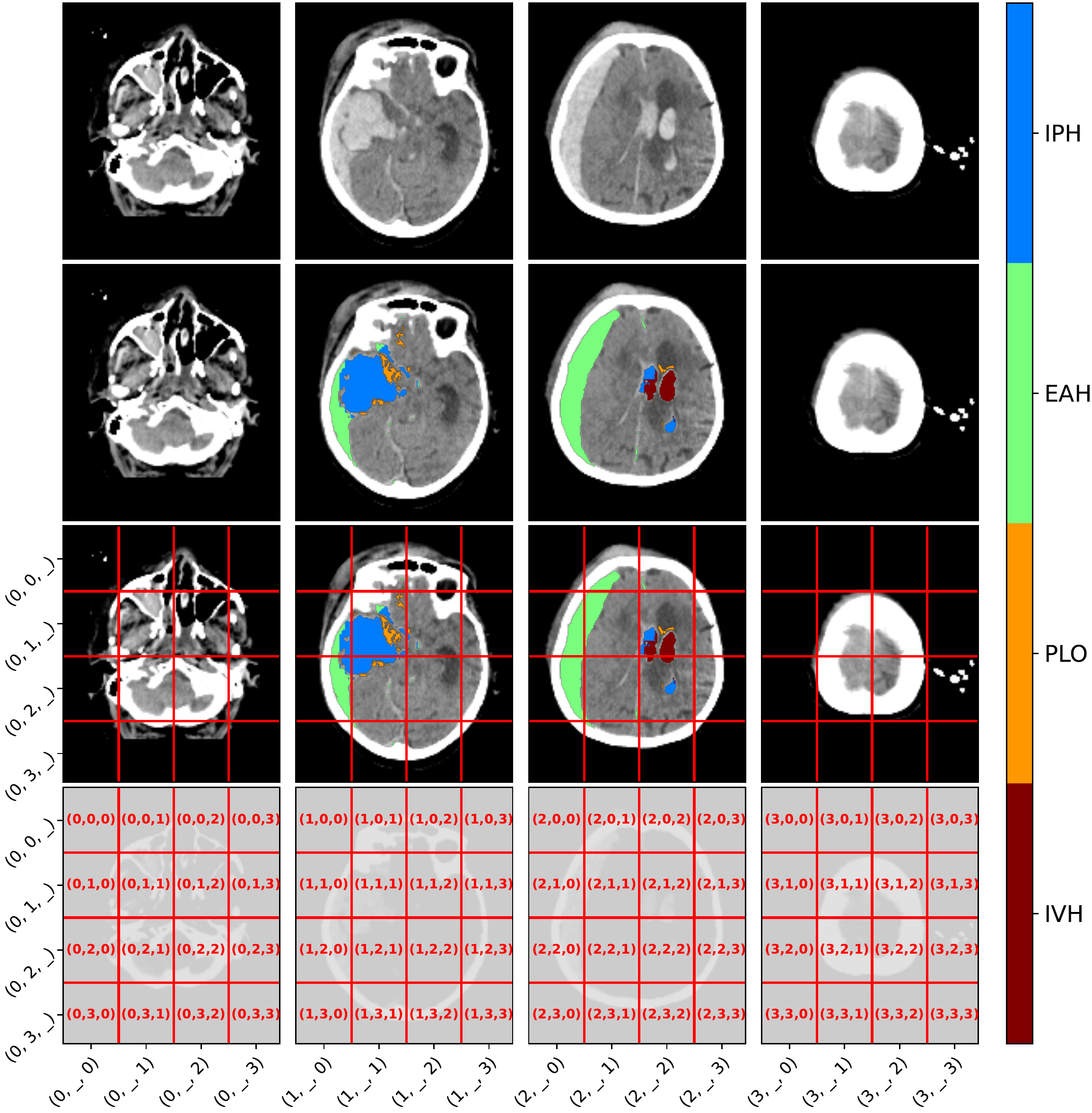}
    \caption{
     Example of feature extraction. A CT scan (1st row) is fed through a segmentation neural network, outputting lesion volumes (2nd row + legend). We either aggregated the lesion volumes globally or calculated the volumes over each cuboidal region to provide localised measures (3rd and 4th rows).
     }
    \label{fig:brain_partition}
\end{figure}

From the 3D segmentations, we extracted two types of statistical features, \emph{global} and \emph{local} lesion biomarkers.
We calculated all connected components for each lesion type and discarded any connected region of 0.02mL or less to remove noise.
We determined this threshold by visual inspection of the scans.
From the separate lesions\footnote{Each lesion is a connected component.}, we extracted the following \emph{global} lesion biomarkers: the total number of lesions, the median lesion volume, the 25th and 75th lesion volume percentiles, and the mean lesion volume.
Next, we partitioned the registered 3D scans four ways for each axis, creating $4^3=64$ equidimensional cuboids. We chose four partitions to balance feature granularity and human interpretability.
We extracted \emph{local} lesion biomarkers for each lesion type by calculating the total lesion volume in each cuboidal region.
Figure~\ref{fig:brain_partition} shows an example of brain partitioning and the corresponding indexing.

\subsubsection*{Modelling and performance evaluation}
We used a Random Forest Classifier with 300 estimators to predict a patient's favourable or unfavourable outcome. An unfavourable outcome, defined as a GOS-E score of 4 or below, is a typical target in TBI outcome prediction. We compared eight predictive models based on different sets of features.
The first four models use imaging features alone: 1) the Marshall score; 2) \emph{global} lesion biomarkers; 3) \emph{local} lesion biomarkers; 4) \emph{global} and \emph{local} lesion biomarkers. The second set of models uses the imaging features above, together with the clinical TBI biomarkers (age, motor and pupil responsiveness, hypoxia and hypotension).
Note that the clinical TBI biomarkers and Marshall score are the same as those used in the state-of-the-art IMPACT-CT model \cite{steyerberg2008predicting}.

We evaluated model performance using the area under the receiver-operator curve (AUROC), precision, recall, and true positive rate at a 10\% false positive rate. In addition to evaluating performance on the holdout test set, we also measured cross-validation performance on the training set.
We calculated statistical significance through a permutation test on the holdout set's metrics and the statistical relevance of each feature using the average Gini importance. In addition, a cross-validation per clinical centre can be found in appendix Table~\ref{tab:app_CV}.

\section{Results}

\begin{table}[!t]
\centering
\caption{Models performance when predicting unfavourable outcome. Marshall score is described in \cite{marshall1992diagnosis}, \emph{global} features refer to disjoint lesion statistics, and \emph{local} features refer to lesion volume per cuboid (see \textbf{Feature extraction}). The clinical biomarkers are age, motor and pupil responsiveness, hypoxia and hypotension.}
\label{tab:res}

\begin{tabular}{|l|c|c|c|c|} 
\hline

\multicolumn{1}{|c|}{\multirow{2}{*}{\textcolor[rgb]{0,0.424,0.686}{\textbf{Features}}}}                                   & \textcolor[rgb]{0,0.424,0.686}{\textbf{Cross-validation}} & \multicolumn{3}{c|}{\textcolor[rgb]{0,0.424,0.686}{\textbf{Hold-out set}}}  \\

                                                                                                     & AUROC                                                     & AUROC                & Precision            & Recall                        \\ 
\hline

Marshall score                                                                                       & 76.7 +/- 7.7                                              & 73.2                 & 69.6                 & 61.5                          \\ 
\hline
\emph{global} lesion features                                                                                    & 72.4 +/- 5.8                                              & 80.9                 & 70.0                 & \textbf{80.8}                 \\ 
\hline
\emph{local} lesion features                                                                                      & 77.2 +/- 5.5                                              & 83.3                 & 65.6                 & \textbf{80.8}                 \\ 
\hline
\textbf{\emph{local} + \emph{global} lesion features}                                                                   & \textbf{77.2 +/- 6.5}                                     & \textbf{84.0}        & \textbf{74.1}        & 76.9                          \\
\hline
\hline
\multicolumn{1}{|c|}{\multirow{2}{*}{\textcolor[rgb]{0,0.424,0.686}{\textbf{Features}}}} & \textcolor[rgb]{0,0.424,0.686}{\textbf{Cross-validation}} & \multicolumn{3}{c|}{\textcolor[rgb]{0,0.424,0.686}{\textbf{Hold-out set}}}  \\
\multicolumn{1}{|c|}{}                                                                   & AUROC                                                     & AUROC         & Precision     & Recall                                      \\ 
\hline
Marshall score~+ clinical biomarkers                                                         & 81.6 +/- 3.9                                              & 84.7          & 69.2          & 69.2                                        \\ 
\hline
\emph{global} + ~clinical biomarkers                                                                 & 82.1 +/- 4.3                                              & 87.5          & 80.8          & \textbf{80.8}                               \\ 
\hline
\textbf{\emph{local} }~\textbf{+ clinical biomarkers}                                                  & \textbf{83.0 +/- 4.4}                                     & \textbf{87.7} & \textbf{84.0} & \textbf{80.8}                               \\ 
\hline
\emph{local}  + \emph{global} + clinical biomarkers                                                           & 81.1 +/- 4.8                                              & 87.2          & 77.8          & \textbf{80.8}                               \\
\hline
\end{tabular}

\end{table}
\begin{figure}[!t]
    \centering
     \begin{subfigure}[b]{.49\textwidth}
         \centering
         \includegraphics[width=\textwidth]{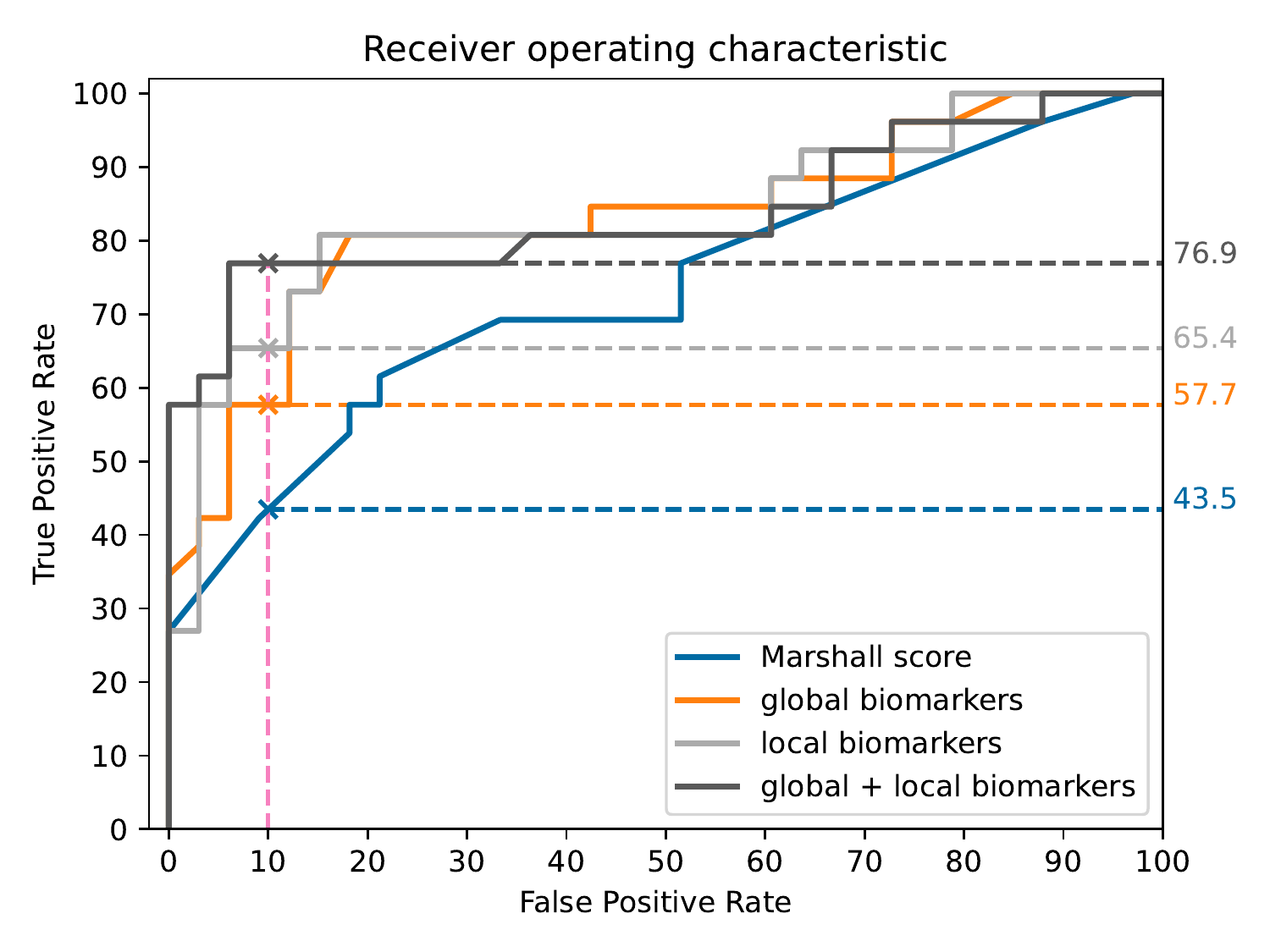} 
         \caption{Imaging biomarkers}
         \label{fig:ROC_curves:img}
     \end{subfigure}
     \hfill
     \begin{subfigure}[b]{.49\textwidth}
         \centering
         \includegraphics[width=\textwidth]{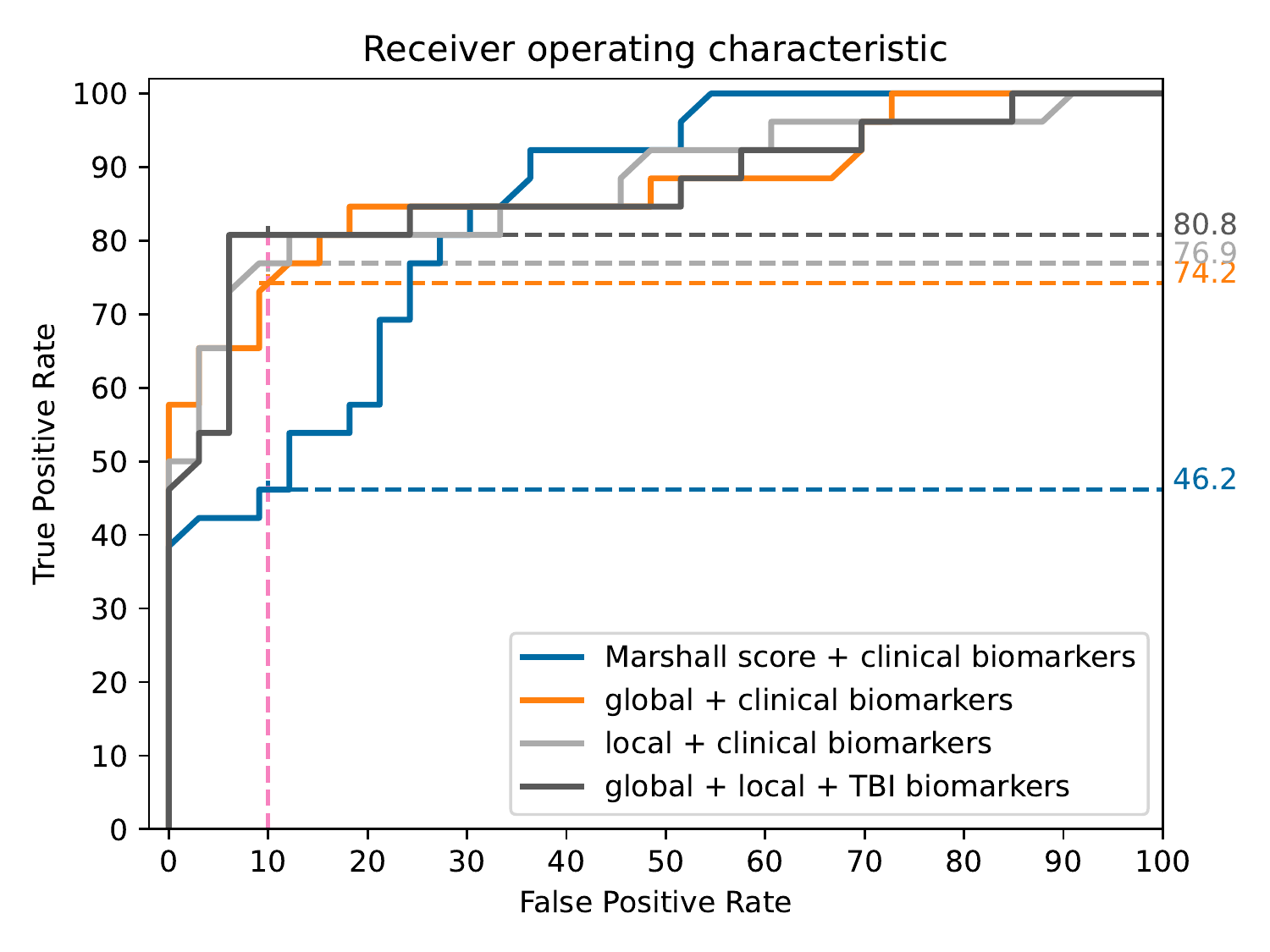}
         \caption{Imaging + clinical biomarkers}
         \label{fig:ROC_curves:img+TBI}
     \end{subfigure}
     \caption{ROC curves of models predicting unfavourable outcome. The dashed lines indicate the true positive rate for a fixed false positive rate of 10\%. \emph{Global} and \emph{local} biomarkers always produce a better or equivalent performance than the Marshall score.}
    \label{fig:ROC_curves}
\end{figure}
\subsubsection*{\emph{Local} and \emph{global} lesion biomarkers performed similarly or better than Marshall score.}
Using \emph{local} and \emph{global} lesion biomarkers achieved a cross-validation AUROC of 76.7 $\pm$ 7.7\% compared to 77.0 $\pm$ 6.6\% when using the Marshall score (Table~\ref{tab:res} top).
On the holdout set, the improvement of AUROC was 10.8\%, from 73.2\% using the Marshall score to 84.0\% using \emph{local} and \emph{global} lesion biomarkers.
Similarly, the precision improved by 4.5\% and the recall by 15.4\%. 
For a false positive rate of 10\%, the volumetric features yielded a true positive rate of 73.1\% compared to 43.5\% for the Marshall score (Figure~\ref{fig:ROC_curves:img}).
Testing the statistical significance of these improvements, we found that improvement in AUROC in the holdout set was statistically significant (one-way p-value $<$ 0.05), whereas all other metrics on the holdout set were statistically comparable.

\subsubsection*{\emph{Local} lesion biomarkers and clinical features performed similarly or better than features used in IMPACT-CT.}
When adding the clinical TBI features to the \emph{local} lesion biomarkers, the AUROC was 83.0 $\pm$ 4.4\% in cross-validation and 87.7\% on the holdout set, compared to 81.6 $\pm$ 3.9\% and 84.7\%, respectively, for clinical TBI and Marshall score biomarkers ((Table~\ref{tab:res}) bottom).
Similarly, the holdout set's precision, recall and true-positive rate improved by 3\%, 14.8\% and 34.6\%, respectively (Figure~\ref{fig:ROC_curves:img+TBI}).
When tested, the improvement in true positive rate was significant (one-way p-value $<$ 0.05), whereas the remaining holdout set metrics for the two experiments were statistically comparable.

\begin{figure}[!t]
    \centering
    \begin{subfigure}[t]{.45\textwidth}
         \centering
         \includegraphics[width=\textwidth]{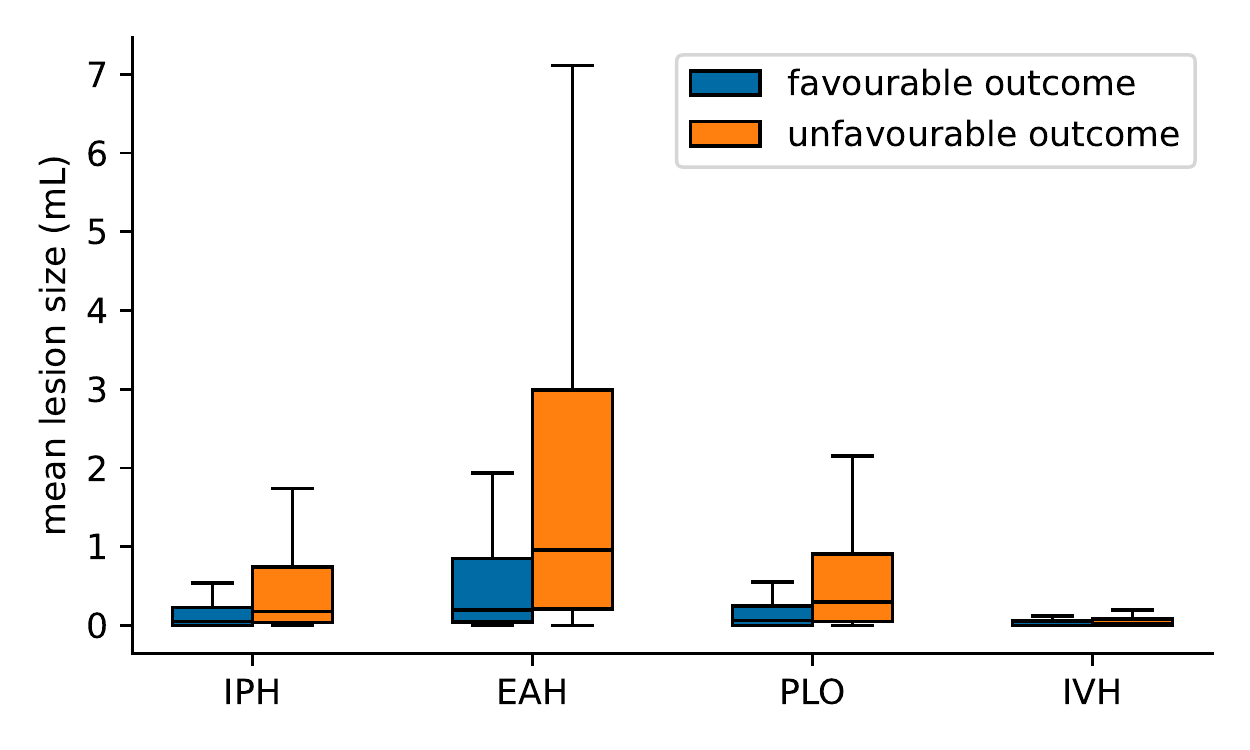}
    \caption{Mean lesion size stratified per lesion type and outcome. Unfavourable outcome can be qualified by larger average IPH, EAH and PLO.}
    \label{fig:boxplot}
    \end{subfigure}
    ~
    \begin{subfigure}[t]{.45\textwidth}
         \centering
    \includegraphics[width=\textwidth]{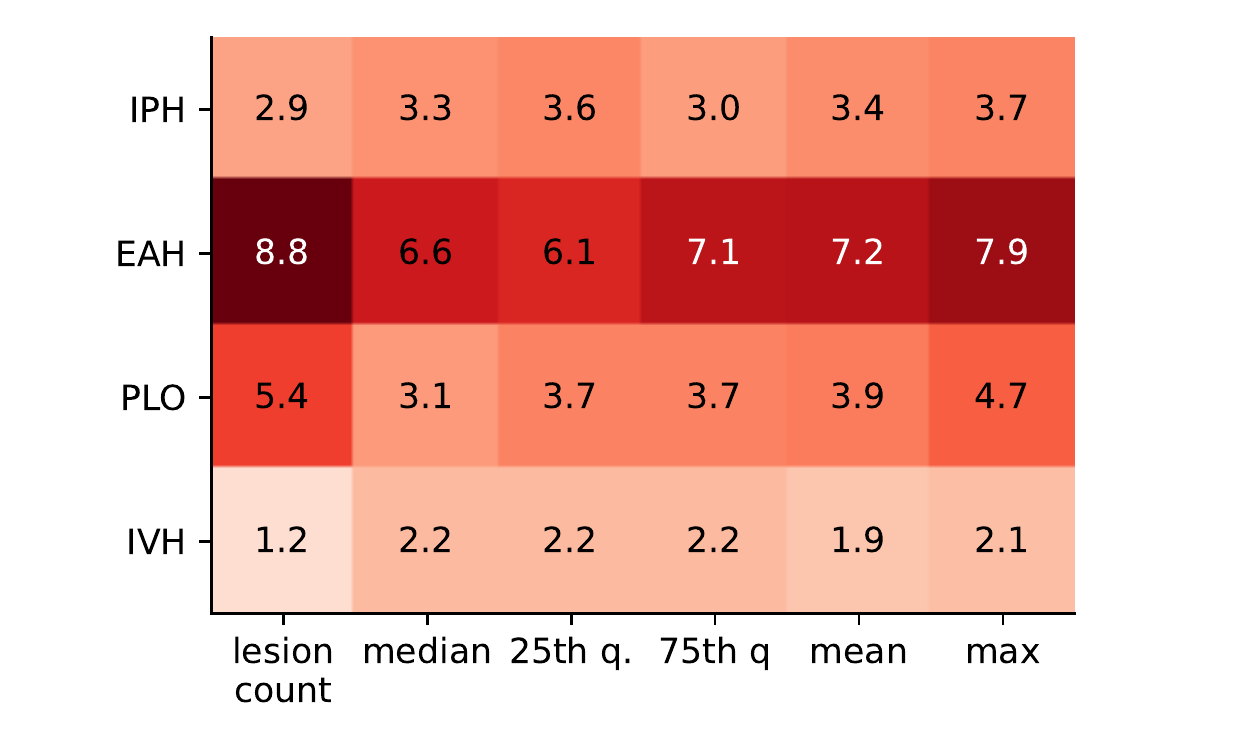}
    \caption{Feature importance of \emph{global} lesion biomarkers when predicting unfavourable outcome. The numbers and colour intensity refer to the Gini importance (rescaled by $10^{-2}$).}
    \label{fig:feat_imp_stat}
    \end{subfigure}
    \caption{\emph{Global} feature statistics and their importance for outcome prediction.}
\end{figure}

\subsubsection*{Extra-axial haemorrhage was the most important feature.}
When considering feature importance when using \emph{global} lesion biomarkers (Figure~\ref{fig:feat_imp_stat}),
EAH was the statistical feature with the highest importance score, and IVH was the feature with the lowest. The lesion count and maximum lesion size were the most important factors.
\begin{figure}[!t]
    \centering
    \includegraphics[width=.75\textwidth]{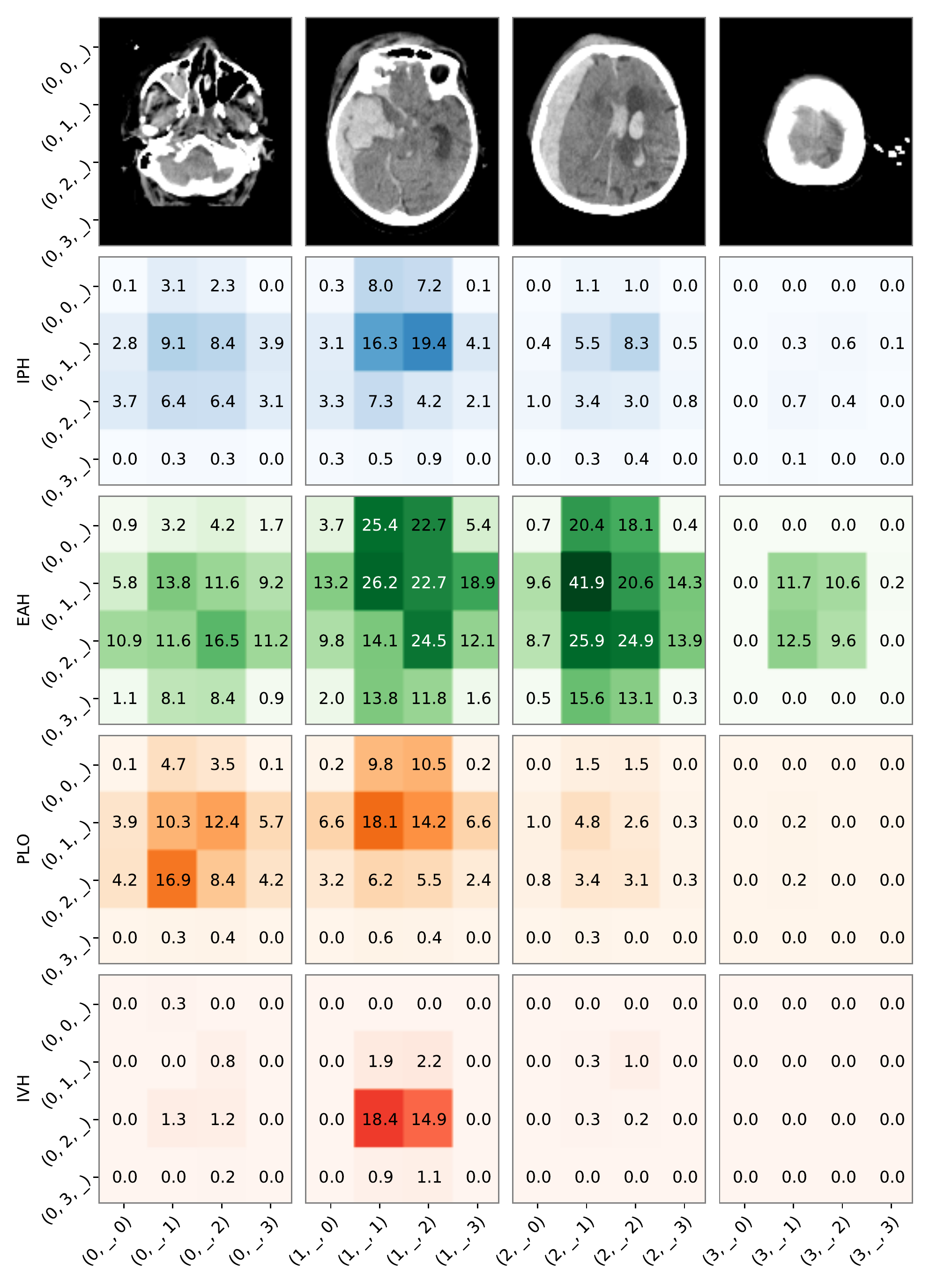}
    \caption{Feature importance of local lesion biomarkers when predicting unfavourable outcome. Each row represents the feature importance of different lesion types, and each column represents a transversal slice, similarly to Figure~\ref{fig:brain_partition}. For each transverse slice, each row represents a coronal slice, whereas each column represents a sagittal slice. The numbers and colour intensity refer to the Gini importance (rescaled by $10^{-3}$).}
    \label{fig:feat_imp_loc}
\end{figure}
As per the \emph{global} lesion biomarkers, EAH was the statistical feature with the highest importance scores when using \emph{local} lesion biomarkers to predict unfavourable outcome (Figure~\ref{fig:feat_imp_loc}). In addition IVH in the second-bottom transverse plane, second-bottom coronal plane ($(1,2,\_)$), EAH in the second-front coronal plane ($(1,1,\_)$ and $(2,1,\_$) were important.

\section{Discussion}

We show that the predictive power of the automatically extracted imaging features is comparable to or superior to that of the Marshall score. Furthermore, the automatically extracted features in conjunction with clinical TBI biomarkers perform at least as well as the features used in IMPACT-CT (Marshall score and clinical TBI biomarkers). The advantage of our approach is that, unlike the Marshall score, automatically extracted imaging features do not require a radiologist to manually review the scan, allowing faster patient care, and reducing workload. 
Our method generalises well across different scanner types and acquisition protocols as shown from the consistent results on the training set cross-validation and the independent hold-out set.
Although other approaches, such as advanced fluid biomarker or magnetic resonance imaging, have also shown promise in improving outcome prediction \cite{graham2021axonal}, the described method has the advantage of using data which is currently collected routinely, obviating the need for revised clinical investigation protocols.

The interpretability of the lesion features is an important step in discovering data-driven prognostic biomarkers, contributing to the clinical understanding of TBI. Reinforcing previous results \cite{atzema2006prevalence}
, we found that frontal EAH is an important indicator of poor TBI outcomes.

Although no ground truth segmentation was available, the clinicians (LL and DS) 
qualitatively reviewed a subset of the automatic segmentations.
We concluded that the segmentation model tended to produce some 
errors, such as partially mislabelling lesion types and under-estimating their size. An example can be seen in the appendix Figure \ref{fig:mis-segmented}. Unfortunately, given the absence of ground truth, we could not quantify the extent of these issues.
Nevertheless, there is strong evidence for the soundness of the model through both the predictive performance and the feature importance maps. For example, the feature importance of IVH is high where the ventricles occur.

In summary, our results show that automatically extracted CT features achieve human-level performance in predicting the outcome of TBI patients without requiring manual appraisal of the scan.
In future work, the interpretability of the machine learning features may allow for a deeper clinical understanding of TBI, a notably complex condition.
However, further work is needed to improve the robustness of lesion segmentation models and their evaluation on new unlabelled datasets.
\FloatBarrier

\subsubsection*{Acknowledgements}
MR is supported by UK Research and Innovation [UKRI Centre for Doctoral Training in AI for Healthcare grant number EP/S023283/1].
LL is supported by NIHR, Academy of Medical Sciences.
NSNG is funded by a National Institute for Health Research (NIHR) Academic Clinical Lectureship and acknowledges infrastructure support from the UK Dementia Research Institute (DRI) Care Research and Technology Centre at Imperial College London and the NIHR Imperial Biomedical Research Centre (BRC).
The BIOAX-TBI data was collected thanks to the ERA-NET NEURON Cofund (MR/R004528/1), part of the European Research Projects on External Insults to the Nervous System call, within the Horizon 2020 funding framework.
The CREACTIVE project has received funding from the European Union Seventh Framework Programme (FP7/2007-2013) under Grant Agreement number 602714.
\
\bibliographystyle{splncs04}
\small{\bibliography{paper7}}

\newpage
\renewcommand\thefigure{A.\Roman{figure}}
\renewcommand\thetable{A.\Roman{figure}}
\section*{Appendix}
\setcounter{figure}{0}  
\setcounter{table}{0}  
\FloatBarrier
\begin{figure}[!h]
    \centering
    \includegraphics[width=.7\textwidth]{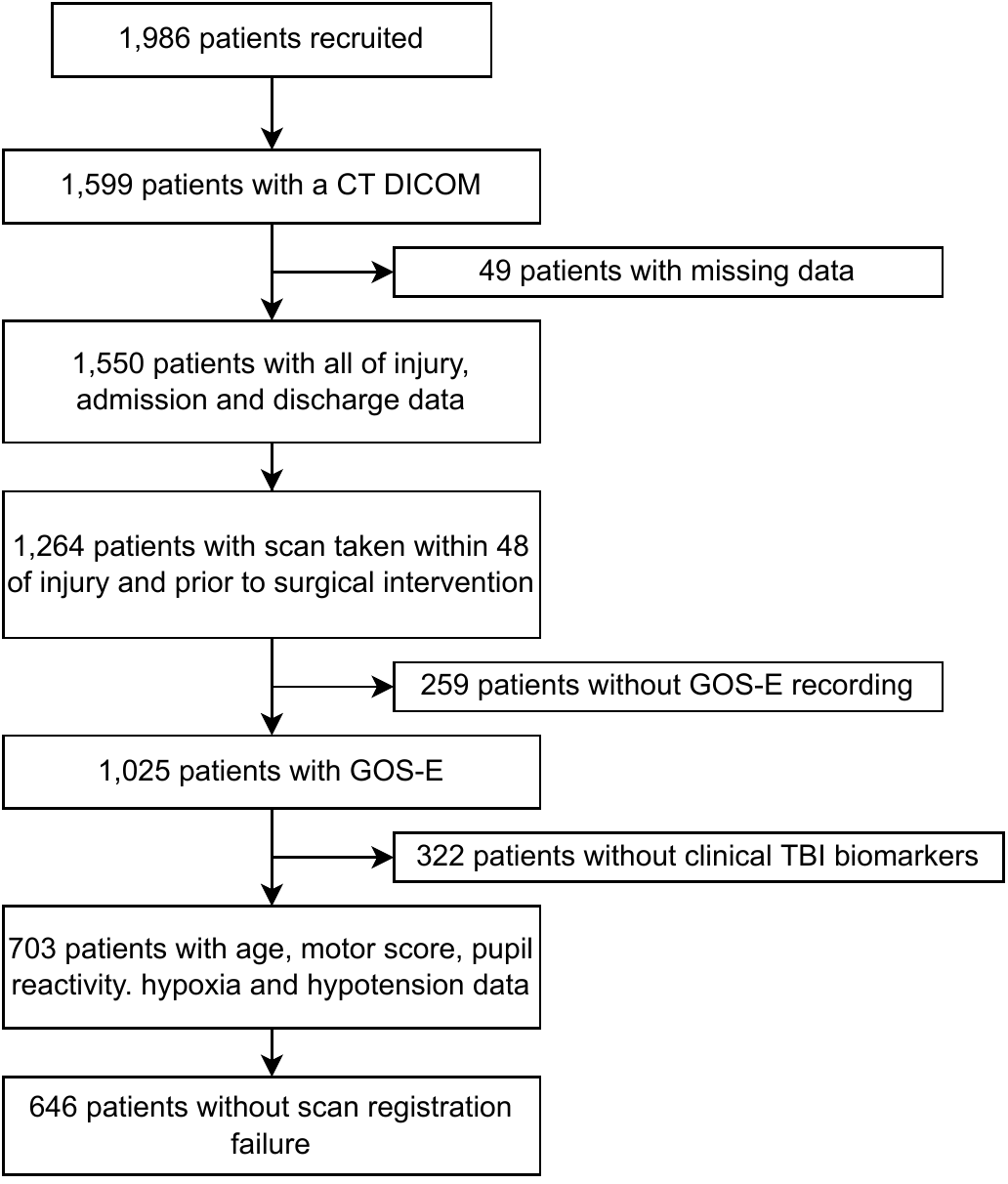}
    \caption{Patient selection flow-chart}
    \label{fig:patient_flow_chart}
    
    \vspace{.5cm}
    \centering
    \includegraphics[width=\textwidth]{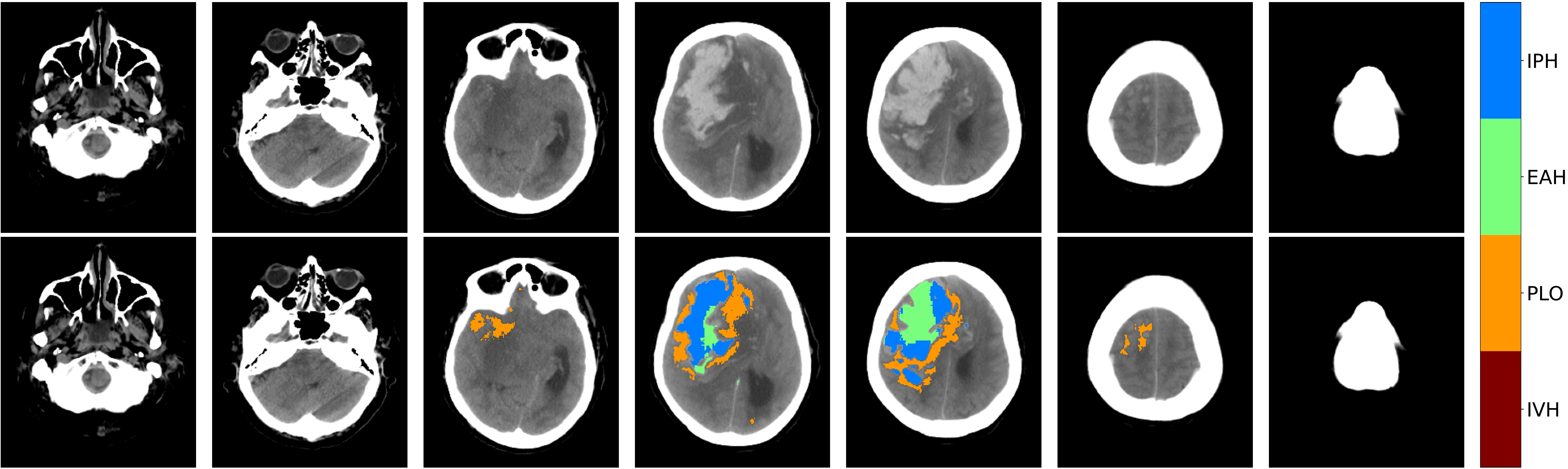}
    \caption{Extreme example of mis-segmented CT scan. The lesion labels are intraparenchymal (IPH), extra-axial (EAH) and intraventricular (IVH) haemorrhages, and perilesional oedemas (PLO). The extra-axial label on fourth and fifth slices is erroneous; the opacity on the left side of the brain in the third slice should be labelled as oedema.}
    \label{fig:mis-segmented}
\end{figure}

\begin{landscape}
\begin{table}
\centering
\caption{Cross validation per centre. Each row represents a trained model, trained on all training data but the centre. The first column is the centre name, the second contains the number of patients, the third the number of patients with unfavourable outcome. The fourth to eleventh columns report the AUROC on the centre data using features described in the first row (for more details see Methods). The results obtained with the Marshall score and global + local features were statistically comparable when taken in isolation and when adding clinical features.}
\label{tab:app_CV}
\arrayrulecolor[rgb]{0.502,0.502,0.502}
\begin{tabular}{!{\color{black}\vrule}c|c|c!{\color{black}\vrule}c|c|c|c!{\color{black}\vrule}c|c|c|c!{\color{black}\vrule}} 
\arrayrulecolor{black}\hline
\textcolor[rgb]{0,0.424,0.686}{Centre}  & \begin{tabular}[c]{@{}c@{}}\textcolor[rgb]{0,0.424,0.686}{N. test patients}\\\textcolor[rgb]{0,0.424,0.686}{(\% of tot)}\end{tabular} & \begin{tabular}[c]{@{}c@{}}\textcolor[rgb]{0,0.424,0.686}{unfav. outcome}\\\textcolor[rgb]{0,0.424,0.686}{(\% of test)}\end{tabular} & \begin{tabular}[c]{@{}c@{}}\textcolor[rgb]{0,0.424,0.686}{Marshall}\\\textcolor[rgb]{0,0.424,0.686}{~score}\end{tabular} & \textcolor[rgb]{0,0.424,0.686}{global} & \textcolor[rgb]{0,0.424,0.686}{local} & \begin{tabular}[c]{@{}c@{}}\textcolor[rgb]{0,0.424,0.686}{global }\\\textcolor[rgb]{0,0.424,0.686}{+ local}\end{tabular} & \begin{tabular}[c]{@{}c@{}}\textcolor[rgb]{0,0.424,0.686}{Marshall}\\\textcolor[rgb]{0,0.424,0.686}{~score + clinical}\end{tabular}   & \begin{tabular}[c]{@{}c@{}}\textcolor[rgb]{0,0.424,0.686}{global}\\\textcolor[rgb]{0,0.424,0.686}{+ clinical}\end{tabular} & \begin{tabular}[c]{@{}c@{}}\textcolor[rgb]{0,0.424,0.686}{local}\\\textcolor[rgb]{0,0.424,0.686}{+ clinical}\end{tabular} & \begin{tabular}[c]{@{}c@{}}\textcolor[rgb]{0,0.424,0.686}{global + local}\\\textcolor[rgb]{0,0.424,0.686}{+ clinical}\end{tabular} \\ 
\hline
\rowcolor[rgb]{0.922,0.922,0.922} SI009 & 80 (12.4\%)             & 80 (56.7\%)            & 66.5       & 62.8           & \textbf{69.5} & 68.3       & \textbf{76.6}           & 73.8   & 74.1  & 72.8 \\
IT079           & 71 (11.0\%)               & 71 (59.2\%)            & \textbf{78.1}                      & 72.0           & 75.8          & 75.1       & \textbf{83.8}           & 83.1   & 82.0  & 78.5   \\
\rowcolor[rgb]{0.922,0.922,0.922} IT100 & 79 (12.2\%)             & 79 (79\%)              & 90.0       & 85.3           & \textbf{90.7} & 89.3       & 90.9           & 93.4   & \textbf{94.0}  & 93.0            \\
IT544           & 30 (4.6\%)              & 30 (51.7\%)            & 77.3       & 74.0           & 77.5          & \textbf{80.4}                      & \textbf{82.8}           & 80.4   & 81.6  & 81.7            \\
\rowcolor[rgb]{0.922,0.922,0.922} IT064 & 27 (4.2\%)              & 27 (64.3\%)            & 83.7       & 70.6           & \textbf{84.1} & 81.1       & 80.1           & 79.5   & \textbf{90.6}  & 85.8            \\
IT442           & 14 (2.2\%)              & 14 (66.7\%)            & 77.6       & \textbf{79.6}  & 62.2          & 67.3                               & \textbf{83.7}           & 69.4   & 75.5  & 71.4            \\
\rowcolor[rgb]{0.922,0.922,0.922} IT062 & 11 (1.7\%)              & 11 (64.7\%)            & \textbf{79.5}             & 59.8   & 66.7   & 57.6       & 81.8           & \textbf{90.2}   & 86.4  & 84.1            \\
IT099           & 11 (1.7\%)              & 11 (64.7\%)            & 63.6       & 62.1           & \textbf{81.8} & 78.8                               & \textbf{89.4}           & 81.8   & 86.4  & 84.8            \\
\rowcolor[rgb]{0.922,0.922,0.922} IT651 & 5 (0.8\%)               & 5 (38.5\%)             & 55.0       & \textbf{85.0}  & 77.5          & 82.5       & 65.0             & \textbf{90.0}   & 85.0  & 87.5            \\
IT513           & 8 (1.2\%)               & 8 (66.7\%)             & \textbf{81.3}                      & 71.9           & 68.8          & 78.1       & 70.3           & \textbf{78.1}   & 75.0  & \textbf{78.1}            \\
\rowcolor[rgb]{0.922,0.922,0.922} IT101 & 3 (0.5\%)               & 3 (37.5\%)      & \textbf{86.7}         & 53.3           & 80.0        & 66.7     & \textbf{93.3}           & 56.7   & 73.3  & 73.3            \\
CH001           & 1 (0.2\%)               & 1 (12.5\%)             & 42.9       & \textbf{71.4}  & 42.9          & 57.1                               & 71.4           & \textbf{85.7}   & 57.1  & 57.1            \\
\rowcolor[rgb]{0.922,0.922,0.922} IT036 & 6 (0.9\%)               & 6 (85.7\%)             & 91.7       & \textbf{100.0} & 83.3          & 66.7       & \textbf{100.0}            & 83.3   & 83.3  & 83.3            \\
IT034           & 4 (0.6\%)               & 4 (57.1\%)             & 95.8       & 91.7           & \textbf{100.0}                   & \textbf{100.0}  & \textbf{100.0}            & \textbf{100.0}  & \textbf{100.0} & \textbf{100.0}           \\
\rowcolor[rgb]{0.922,0.922,0.922} IT590 & 3 (0.5\%)               & 3 (50\%)               & 83.3       & 77.8     & \textbf{100.0}  & \textbf{100.0} & 55.6           & 77.8   & \textbf{100.0} & \textbf{100.0}           \\
IT724           & 4 (0.6\%)               & 4 (100\%)              & na         & na             & na            & na         & na             & na     & na    & na              \\
\rowcolor[rgb]{0.922,0.922,0.922} IT057 & 4 (0.6\%)               & 4 (100\%)              & na         & na             & na            & na         & na             & na     & na    & na              \\
IT088           & 2 (0.3\%)               & 2 (100\%)              & na         & na             & na            & na         & na             & na     & na    & na              \\
\hline
\end{tabular}
\end{table}
\end{landscape}

\end{document}